%% file: naaclhlt2019.tex
\documentclass[11pt,a4paper]{article}
\usepackage[hyperref]{naaclhlt2019}
\usepackage{times}
\usepackage{latexsym}

\usepackage{url}
\usepackage{graphicx}  %Required
\usepackage{amsmath}
\usepackage{algorithm}
\usepackage{algpseudocode}
\usepackage{amsfonts}
\usepackage{amssymb}
\usepackage{bm}
\usepackage{color}
\usepackage{pgfplots}
\usepackage{subcaption}
\usepackage{float}

\newcommand{\vD}{{\bm{D}}}
\newcommand{\vx}{{\bm{x}}}

\newcommand{\vy}{{\bm{y}}}
\newcommand{\vz}{{\bm{z}}}

\newcommand{\vtheta}{{\bm{\theta}}}

\newcommand{\softmax}{\textrm{softmax}}
\newcommand{\sparsemax}{\textrm{sparsemax}}

\newcommand{\sameen}[1]{{\color{black} #1}}

\newenvironment{itemizesquish}{\begin{list}{\labelitemi}{\setlength{\itemsep}{0em}\setlength{\labelwidth}{0.5em}\setlength{\leftmargin}{\labelwidth}\addtolength{\leftmargin}{\labelsep}}}{\end{list}}
\newenvironment{enumeratesquish}{\begin{list}{\addtocounter{enumi}{1}\labelenumi}{\setlength{\itemsep}{0em}\setlength{\labelwidth}{0.5em}\setlength{\leftmargin}{\labelwidth}\addtolength{\leftmargin}{\labelsep}}}{\end{list}\setcounter{enumi}{0}}

\aclfinalcopy % Uncomment this line for the final submission
\title{Selective Attention for Context-aware Neural Machine Translation}

\author{Sameen Maruf$^{\dagger}$\thanks{$^*$Work initiated during an internship at Unbabel.} \\\And
  Andr{\'e} F. T. Martins$^{\ddagger}$ \\
  $^{\dagger}$Faculty of Information Technology, Monash University, VIC, Australia\\
  $^{\ddagger}$Unbabel \& Instituto de Telecomunica\c{c}\~oes, Lisbon, Portugal\\
  {\tt {$^{\dagger}$\{firstname.lastname\}@monash.edu}} \\
  {\tt $^{\ddagger}$andre.martins@unbabel.com} \\\And
  Gholamreza Haffari$^{\dagger}$\\}

\date{}

\begin{document}
\maketitle
\begin{abstract}
Despite the progress made in sentence-level NMT, current systems still fall short at achieving fluent, good quality translation for a full document. Recent works in context-aware NMT consider only a few previous sentences as context and may not scale to entire documents. To this end, we propose a novel and scalable top-down approach to hierarchical attention for context-aware NMT which uses sparse attention to selectively focus on relevant sentences in the document context and then attends to key words in those sentences. We also propose single-level attention approaches based on sentence or word-level information in the context. The document-level context representation, produced from these attention modules, is integrated into the encoder or decoder of the Transformer model depending on whether we use monolingual or bilingual context. Our experiments and evaluation on English-German datasets in different document MT settings show that our selective attention approach not only significantly outperforms context-agnostic baselines but also surpasses context-aware baselines in most cases. 
%\andre{Should we mention selective attention (and connection to sparsity) in the abstract? If we do some analysis that show interpretablity, we can claim here benefits in terms of interpretability as well. We need to make this abstract sound more exciting :)}
\end{abstract}

\section{Introduction}
\input{sec1-intro.tex}

\section{Background}
\input{sec3-preliminaries.tex}

\section{Proposed Approach}
\input{sec4-model.tex}

\section{Experiments}
\input{sec5-setup.tex}

%\section{Results}
\input{sec6-results.tex}

\section{Related Work}
\input{sec2-related-work.tex}

\section{Conclusion}
\input{sec8-conclusions.tex}

\section*{Acknowledgments}
The authors are grateful to the anonymous reviewers for their helpful comments and corrections. SM would like to thank her colleagues at Monash University: Veronika Kuchta, Harald B{\"o}geholz and Hagen Lauer, for their help in the subjective evaluation. This work was supported by the Multi-modal Australian ScienceS Imaging and Visualisation Environment (MASSIVE) (\url{www.massive.org.au}), 
the European Research Council (ERC StG DeepSPIN 758969),
 the Funda\c{c}\~ao para a Ci\^encia e Tecnologia through contracts UID/EEA/50008/2019 and CMUPERI/TIC/0046/2014 (GoLocal), and 
 a Google Faculty Research Award to GH.

\bibliography{naaclhlt2019}
\bibliographystyle{acl_natbib}

%\appendix

\end{document}

%% file: sec1-intro.tex
Neural machine translation has grown immensely in the past few years, from the simplistic RNN-based encoder-decoder models \cite{Sutskever:14, Bahdanau:15} to the state-of-the-art Transformer architecture \cite{Vaswani:17}. Most of these models rely on the attention mechanism as a major component, which involves focusing on different parts of a sequence to compute new representations, and has proven to be quite effective in improving the translation quality \cite{Vaswani:17}. However, all of these models share the same inherent problem: the translation is still performed on a sentence-by-sentence basis, thus ignoring the long-range dependencies which may be useful when it comes to translating discourse phenomena. 

More recently, {\bf context-aware NMT} has been gaining significant traction from the MT community with majority of works coming out in the past two years. Most of these focus on using a few previous sentences as context \cite{Jean:17, Wang:17, Tu:18, Voita:18, Zhang:18, Miculicich:18} and neglect the rest of the document. Only one existing work has endeavoured to consider the full document context \cite{Maruf:18}, thus proposing a more generalised approach to document-level NMT.  
%\andre{I think you should also cite your WMT18 paper here, maybe as ``an application of context-aware NMT for translating online conversations.'' If not here, at least in the related work section.} 
%They model the sentence-level information in the document implicitly using memory networks \cite{Sukhbaatar:15}. 
However, the model is restrictive as the document-level attention computed is sentence-based and static (computed only once for the sentence being translated). A more recent work \cite{Miculicich:18} proposes to use a hierarchical attention network (HAN) \cite{Yang:16} to model the contextual information in a structured manner using word-level and sentence-level abstractions; yet, it uses a limited number of past source and target sentences as context and is not scalable to entire document. 

In this work, we propose a {\bf selective attention} approach to first selectively focus on relevant sentences in the global document-context and then attend to key words in those sentences, while ignoring the rest.\footnote{The term ``selective attention'' comes from cognitive science and is defined as the act of focusing on a particular object for a period of time while simultaneously ignoring irrelevant information that is also occurring \cite{Dayan:00}.} 
Towards this goal, we use {\bf sparse attention}, enabling an efficient and scalable use of the context. The intuition behind this is the way humans translate a sentence containing ambiguous words. They may look for sentences in the whole document which contain similar words and just focus on those for the translation. This attention, which we call Hierarchical Attention, is computed dynamically for each query word. Furthermore, we propose a Flat Attention approach which is based on either sentence or word-level information in the context. We integrate the document-level context representation, produced from these attention modules, into the encoder or decoder of the Transformer model depending on whether we consider monolingual (source-side) or bilingual (both source and target-side) context.

Our  contributions are as follows: (i) we propose a novel and efficient top-down approach to hierarchical attention for context-aware NMT, (ii) we compare variants of selective attention with both context-agnostic and context-aware baselines, and (iii) we run experiments in both online (only past context) and offline (both past and future context) settings on three English-German datasets.
Experiments show that our approach improves upon the Transformer by an overall +1.34, +2.06 and +1.18 BLEU for TED Talks, News-Commentary and Europarl, respectively. It also outperforms two recent context-aware baselines \cite{Zhang:18, Miculicich:18} in majority of the cases.

%% file: sec3-preliminaries.tex
\subsection{Neural Machine Translation}

Generic NMT models are based on an encoder-decoder architecture \cite{Bahdanau:15, Vaswani:17}. The encoder reads the source sentence denoted by $\vx$ =  ($x_1$, $x_2$, ..., $x_M$) and maps it to a continuous representation $\vz$ = ($z_1$, $z_2$, ..., $z_M$). Given $\vz$, an attentional decoder generates the target translation $\vy$ = ($y_1$, $y_2$, ..., $y_N$) one word at a time in a left-to-right fashion.
The popular Transformer architecture \cite{Vaswani:17} follows the same structure by using stacked self-attention and point-wise, fully connected layers for both the encoder and decoder.

\paragraph{Encoder}
The encoder stack is composed of $L$ identical layers, each containing two sub-layers. The first, a multi-head self-attention sub-layer, %and the second is a simple, position-wise fully connected feed-forward network. The self-attention layer %all of the keys, values and queries come from the same place, i.e, the output of the previous layer in the encoder. Thus, 
allows each position in the encoder to attend to all positions in the previous layer of the encoder, while the second, a feed-forward network, uses two linear transformations with a ReLU activation.

\paragraph{Decoder}
The decoder stack is also composed of $L$ identical layers. In addition to the two sub-layers, the decoder inserts a third sub-layer, which performs multi-head attention over the output of the encoder stack. Masking is used in the self-attention sub-layer %in the decoder stack %is also modified 
to prevent positions from attending to subsequent positions thus avoiding leftward flow of information.

\subsection{Document-level Machine Translation}\label{sec:docmt}

In general, the probability of a document translation $\bm{Y}$ given the source document $\bm{X}$ is given by:
%\vspace{-1mm}
\begin{equation}\label{eq:energy}
P_{\vtheta}(\bm{Y}|\bm{X})=\prod_{j=1}^{J} P_{\vtheta}(\vy^j|\vx^j, \bm{D}_{-j})
\end{equation}
where $\vy^j$ and $\vx^j$ denote the $j^{th}$ target and source sentence, respectively, and $\bm{D}_{-j}=\{\bm{X}_{-j},\bm{Y}_{-j}\}$ 
is the collection of all other sentences in the source and target document. Since generic NMT models translate one word at a time, Eq.~\ref{eq:energy} becomes:
\begin{equation}\label{eq:objective}
P_{\vtheta}(\bm{Y}|\bm{X}) = \prod_{j=1}^{J}\prod_{n=1}^{N} P_{\vtheta}(y_{n}^j |\vy_{<n}^j, \vx^j, \bm{D}_{-j}) 
\end{equation}
where $y_{n}^j$ is the $n^{th}$ word of the $j^{th}$ target sentence and $\vy_{<n}^j$ are the previously generated words. 

%Document MT can be divided into two categories depending on the conditioning context $\vD$ if from both past and future (referred to as \textit{offline} setting) or only from the past (referred to as \textit{online} setting). 

\paragraph{Training}
The document-conditioned NMT model $P_{\vtheta}(\vy^j|\vx^j, \bm{D}_{-j})$ is realised using a neural architecture and usually trained via a two-step procedure \cite{Maruf:18, Miculicich:18}. The first step involves pre-training a standard sentence-level NMT model, and the second step involves optimising the parameters of the whole model, i.e., both the document-level and the sentence-level parameters. 
%A drawback of these approaches is that they require visiting each sentence multiple times, regardless of how relevant it is to the rest of the document. Our approach, to be described shortly, sidesteps this problem by enabling a more efficient use of memory.

\paragraph{Decoding}
To generate the best translation for a full document according to the document MT model, the 
problem of maximizing Eq.~\ref{eq:energy} 
%optimisation problem:
%%\vspace{-1mm}
%\begin{equation}
%\arg\max_{\vy^1,\ldots,\vy^{J}} \prod_{j=1}^{J} P_{\vtheta}(\vy^j|\vx^j, \bm{D}_{-j}),
%\end{equation}
is solved using a two-pass Iterative Decoding strategy \cite{Maruf:18}: first, the translation of each sentence is initialised using the sentence-based NMT model; then, each translation is updated using the context-aware NMT model fixing the other sentences' translations.

%% file: sec4-model.tex
The main goal of this paper is to have a document-level NMT model which is memory-efficient, scalable, and capable of listening to the entire document. To achieve this, we augment a sentence-level NMT model (the Transformer \cite{Vaswani:17}) with an efficient hierarchical attention mechanism which has the ability to identify the key sentences in the document context and then attend to the key words within those sentences. As mentioned previously, we want to maximise $P_{\vtheta}(\vy^j|\vx^j, \vD_{-j})$, where we take $\vD_{-j}$ to be either the monolingual source or bilingual source and target-side context in two settings: \textit{offline}---the context comes from both past and future, and \textit{online}---the context comes from only the past. 

In this section, we show how to represent the document-level context using our Context Layer, how to regulate the information at the sentence and document-level using context gating and finally we present our integrated model.

\subsection{Document-level Context Layer}
The context $\vD_{-j}$ is modeled via a single Document-level Context Layer comprising of two sub-layers: (i) a Multi-Head Context Attention sub-layer, and (ii) a Feed-Forward sub-layer, where the former consists of either a top-down Hierarchical Attention module or a Flat Attention module (explained shortly), \sameen{and the latter is similar to the Feed-Forward network in the original Transformer architecture}. Each sub-layer is followed by a layer normalisation.%
\footnote{We do not have residual connections after sub-layers in our Document-level Context Layer as we found them to have a deteriorating effect on the translation scores (also reported by \newcite{Zhang:18}).}

\sameen{Let us now describe the attention modules which independently form the Multi-Head Context Attention sub-layer.}

\subsubsection{Hierarchical Attention}

Our hierachical attention module {\textrm{H-Attention}}($Q_s$, $Q_w$, $K_s$, $K_w$, $V_w$) (Figure~\ref{fig:hieratt}) is a reformulation of the Scaled Dot-Product Attention of \newcite{Vaswani:17}. Here, we have five inputs consisting of two types of keys and queries, one each for the sentences and the words, while the values are based only on words in the context. The Hierarchical Attention module has four operations:

\begin{figure}[t]
 \centering
  \includegraphics[width=0.34\textwidth]{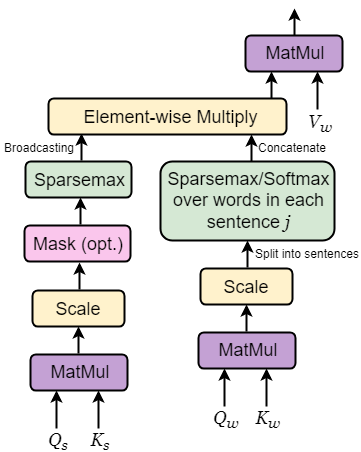}
  \caption{Hierarchical Context Attention module.}
 \label{fig:hieratt}
\end{figure}

\begin{enumeratesquish}
\item \textbf{Sentence-level Key Matching:} This is performed on a set of queries simultaneously, packed together into a matrix $Q_s$. The sentence-level keys are also packed into a matrix $K_s$. We will describe in \S\ref{sec:integrate} how $Q_s$ and $K_s$ are computed. The attention weights are computed as:
\begin{equation}\label{eq:sentence_attention}
\alpha_s = \sparsemax({Q_s{K_s}^T}/{\sqrt{d_k}})
\end{equation}
where $d_k$ is the dimension of the keys, and $\alpha_s$ has dimensions equal to the total number of sentences in the document. \sameen{We propose to use \textbf{sparsemax} \cite{Martins:16}, instead of softmax, as this gives us the intended selective attention behavior, that is identifying the key sentences that may potentially be relevant to the current sentence, hence making the model more efficient in compressing its memory.} A softmax attention, on the other hand, can still assign low probability to sentences, forming a long-tail and absorbing significant probability mass, and it cannot fully {\it ignore} those sentences. 
An additive mask is used (before the \textit{sparsemax} operation) based on whether we train for offline or online setting by masking out only the current sentence or current and future sentences, respectively.

\item \textbf{Word-level Key Matching:} Here the query and key matrices, $Q_w$ and $K_w$, are word-level. We perform a word-level key matching for each sentence $j$ in the document:
\begin{equation}\label{eq:word_attention}
\alpha_{w}^j = \sparsemax({Q_{w}{K_{w}^j}^T}/{\sqrt{d_k}})
\end{equation}
where $\alpha_{w}^j$ is the word-level attention vector for $j^{th}$ sentence.%
\footnote{This can be done for only the sentences with non-zero probabilities (obtained from the sentence-level key matching), however, we found it to be computationally expensive, as it required breaking down the batched matrices.} %
We can also use softmax, instead of sparsemax, for a coarser key matching. We explore the two variants in our experiments.

\item \textbf{Re-scaling attention weights:} The word-level attention is further re-weighted by the corresponding sentence-level attention \cite{Nallapati:16} such that the probability of $j^{th}$ sentence in a document is given by:
\vspace{-1mm}
\begin{equation}
\alpha_{hier}^j = \alpha_s(j)\alpha_{w}^j
\end{equation}
where $\alpha_s(j)$ is the attention weight for the $j^{th}$ sentence obtained via Eq.~\ref{eq:sentence_attention} and $\alpha_w^j$ is as in Eq.~\ref{eq:word_attention}. The re-weighting, thus, produces a scaled attention vector \sameen{$\alpha_{hier}=\textrm{Concat}(\alpha_{hier}^1,...,\alpha_{hier}^J)$}, each entry of which corresponds to the attention weight for a specific word in the document.

\item \textbf{Value Reading:} The set of word-level values is packed together into a matrix $V_w$ and the matrix of outputs is given by ${\alpha_{hier}}V_w$. 
This multiplication, combined with sparsemax attention, allows to {\it prune} the hierarchy. 

\end{enumeratesquish}

We further extend the {\sc MultiHead} attention function proposed by \newcite{Vaswani:17} for our Hierarchical Attention module as: 
%\vspace{0.05cm}
\begin{eqnarray}
& {\textsc{H-MultiHead}}(Q_s, K_s, Q_w, K_w, V_w) = \nonumber \\
& \qquad\qquad\textrm{Concat}(head_1,...,head_H)W^O \nonumber
\end{eqnarray}
where $head_h = {\textrm{H-Attention}}(Q_sW_h^{Q_s}, Q_wW_h^{Q_w}, \\
K_sW_h^{K_s}, K_wW_h^{K_w}, V_wW_h^{V_w})$, $W$'s are parameter matrices and all (five) inputs are transformed using separate linear layers.

\subsubsection{Flat Attention}

%Our hierarchical attention module has the ability to attend to the words in the context in a coarse-to-fine manner. However, 
Another way to model the context $\vD_{-j}$ is via single-level attention by re-using the Scaled Dot-Product Attention in \newcite{Vaswani:17},
\begin{equation}
\textrm{Attention}(Q, K, V) = \softmax({QK^T}/{\sqrt{d_k}})V
\end{equation}
The attention\footnote{We plan to investigate sparse flat attention in future work.} here is of two types: (i) \textit{sentence-level} if $K$, $V$ are computed for sentences in the document, or (ii) \textit{word-level} if $K$, $V$  are computed for words in the document. The former module is similar to the Memory Networks architecture of \newcite{Maruf:18} in that it uses sentence-level information. However, there are two key differences: (i) we use {MultiHead} attention as in the Transformer architecture, and (ii) our context attention is dynamic such that we have a separate attention for each query word.%, while their attention mechanism is static as it is computed once for the whole sentence. 
%\andre{The hierarchical attention model will hopefully be better than this :)}

\subsection{Context Gating}

As mentioned previously, the Multi-Head Context Attention sub-layer is part of the Context Layer (Figure~\ref{fig:hieratt-src}), the output of which is fed into the Transformer architecture through context gating \cite{Tu:18}. For $i^{th}$ word in source or target:
\vspace{-1mm}
\begin{eqnarray}
\gamma_{i}&=&\sigma(W_rr_i+W_dd_i)\\
\tilde{r}_{i}&=&{\gamma}_{i}\odot{r}_{i}+(1-\gamma_{i})\odot{d}_{i}
\end{eqnarray}
where W's are parameter matrices, $r_i$ is the output of encoder or decoder stack for $i^{th}$ word, $d_i$ is the output from the context layer for $i^{th}$ word and $\tilde{r}_{i}$ is the final hidden representation for the same.

\subsection{Integrated Model}\label{sec:integrate}

The context can be integrated into the encoder or decoder of the NMT model depending on if it is monolingual or bilingual.\footnote{\sameen{We do not integrate context into both encoder and decoder as it would have redundant information from the source (the context incorporated in the decoder is bilingual), in addition to increasing the complexity of the model.}}

\begin{figure}[t]
  \centering
 \includegraphics[width=.4\textwidth]{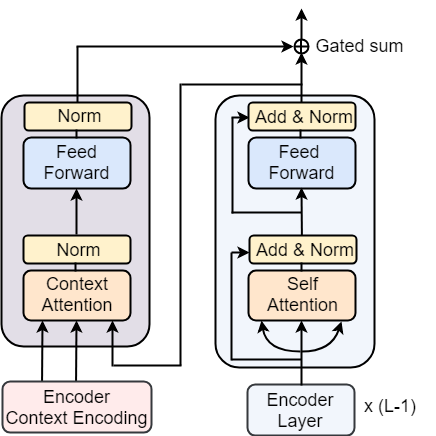}
 \caption{Encoder-side context integration.}
 \label{fig:hieratt-src}
\end{figure}%
\vspace{-1mm}
\begin{figure}[t]
  \centering
  \includegraphics[width=.4\textwidth]{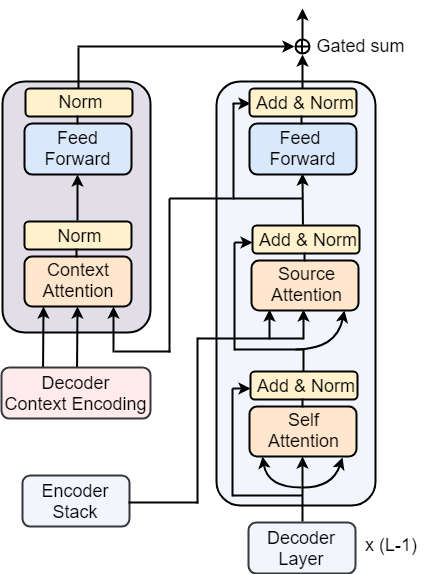}
 \caption{Decoder-side context integration.}
 \label{fig:hieratt-tgt}
\end{figure}

\paragraph{Monolingual context integration in Encoder}
We add the Document-level Context Layer alongside the encoder stack as shown in Figure~\ref{fig:hieratt-src}. The Encoder Context Encoding block stores the keys and values produced from the pre-trained sentence-level NMT model. For word-level attention, the keys $K_w$ and values $V_w$ are composed of vector representations (from last encoder layer) of source words in the document, while for the sentence-level attention, the keys $K_s$ and values $V_s$ are composed of vector representations of sentences in the document where the vector representation of each sentence is an average of the word representations in that sentence. \sameen{The queries $Q_w$, $Q_s$ are linear transformations of the output of the $L^{th}$ encoder layer} which are then matched with the corresponding keys and values stored in the Encoder Context Encoding block just described. 

\paragraph{Bilingual context integration in Decoder}
We again add the Document-level Context Layer alongside the decoder stack as in Figure~\ref{fig:hieratt-tgt}. However, instead of choosing the keys and values to be monolingual as in the encoder, we follow \newcite{Tu:18} in choosing the key to match to the source-side context, while designing the value to match to the target-side context. Hence, the keys (in the Decoder Context Encoding block) are composed of context vectors from the Source Attention sub-layer, while the values are composed of the hidden representations of the target words, both from the last decoder layer. Again the keys $K_w$ and $K_s$ are either for individual target words or target sentences, % in the document, 
and same goes for $V_w$ and $V_s$. The queries $Q_w$, $Q_s$ for the Context Layer come from the Source Attention sub-layer in the $L^{th}$ layer of the decoder (Figure~\ref{fig:hieratt-tgt}).

%% file: sec5-setup.tex
\subsection{Setup}

\paragraph{Datasets}

We conduct experiments for English$\rightarrow$German on three different domains: TED talks, News-Commentary and Europarl. These datasets are chosen based on their variance in genre, style and level of formality:

\begin{itemizesquish}
\item \textbf{TED} This corpus is from the IWSLT 2017 MT track \cite{Cettolo:12} and contains transcripts of TED talks aligned at sentence level. Each talk is considered to be a document. We combine \textit{tst2016-2017} into the test set and the rest are used for development.
\item \textbf{News-Commentary} We obtain the sentence-aligned document-delimited News Commentary v11 corpus for training.\footnote{\scriptsize\url{www.casmacat.eu/corpus/news-commentary.html}} The WMT'16 \textit{news-test2015} and \textit{news-test2016} are used for development and testing, respectively.
\item \textbf{Europarl} This dataset is extracted from Europarl v7 \cite{Koehn:05}. The source and target sentences are aligned using the links provided by \newcite{Tiedemann:12}. Following \newcite{Maruf:18}, we use the \textit{SPEAKER} tag as the document delimiter. Documents longer than 5 sentences are kept and the resulting corpus is randomly split into training, dev and test sets.
\end{itemizesquish}

\input{table-dataset.tex}

The corpora statistics are provided in Table~\ref{table:corpora}. All datasets\footnote{The data is available at \scriptsize\url{https://github.com/sameenmaruf/selective-attn}.} are tokenised and truecased using the Moses toolkit \cite{Koehn:07}, and split into subword units using a joint BPE model with 30K merge operations \cite{Sennrich:16}.

\input{table-offline-results.tex}

\input{table-online-results.tex}

\paragraph{Models and Baselines}

For offline document MT, we have two context-agnostic baselines: (i) a modified version of RNNSearch \cite{Bahdanau:15}, which incorporates dropout on the output layer and improves the attention model by feeding the previously generated word, and (ii) the state-of-the-art Transformer architecture. For the online case, we again have the Transformer as a context-agnostic baseline and two context-aware baselines \cite{Zhang:18, Miculicich:18}.

All models are implemented in C++ using DyNet \cite{dynet}. For RNNSearch, we modify the sentence-based NMT implementation in \textit{mantis} \cite{Cohn:16}. The encoder is a single layer bidirectional GRU \cite{Cho:14} and the decoder is a 2-layer GRU with embeddings and hidden dimensions set to 512. The dropout rate for the output layer is set to 0.2. For the Transformer, we use \textit{Transformer-DyNet}\footnote{\scriptsize\url{https://github.com/duyvuleo/Transformer-DyNet}} implementation and extend it for our context-aware NMT model.%\footnote{The code is available at \scriptsize\url{https://github.com/sameenmaruf/selective-attn}} 
%\andre{should we say somewhere that our code will be made publicly available upon acceptance (Sameen, is this your plan?)} 
The hidden dimensions and feed-forward layer size is set to 512 and 2048 respectively. We use 4 layers\footnote{We found this configuration to be much more stable than using 6 layers with almost no difference in performance as reported by \newcite{Xia:18}.} each in the encoder and decoder with 8 attention heads and employ label smoothing with a value of 0.1. We also employ all four types of dropouts as in the original Transformer with a rate of 0.1 for the sentence-based model and 0.2 for our context-aware model. 

For training all models, we use the default Adam optimiser \cite{Kingma:14} with an initial learning rate of 0.0001 and employ early stopping. For our context-aware NMT model, we use a two-stage training strategy as described in \S\ref{sec:docmt}. For inference, we use Iterative Decoding only when using the bilingual context. All experiments are run on a single Nvidia P100 GPU with 16GBs of memory.\footnote{\sameen{The experiments can also be run on GPUs with 10-12GBs of memory by reducing the batch size at the expense of increased computational cost.}}

\paragraph{Evaluation Metrics}

For evaluation, we use BLEU \cite{Papineni:02} and Meteor \cite{Lavie:07} scores on tokenised text, and measure statistical significance with respect to the baselines, \textit{p} $<$ 0.05 \cite{Clark:11}.

%% file: table-dataset.tex
\setlength{\tabcolsep}{5pt}
\begin{table}[t]
\centering
{\small
\begin{tabular}{l|c|c}
%& & & \multicolumn{2}{c}{Doc Length}\\
%\cline{4-5}
%\textbf{Domain} & \textbf{\#Documents} & \textbf{\#Sentences} & \textbf{Document length}\\
\textbf{Domain} & \textbf{\#Sentences} & \textbf{Document length}\\
\hline \hline
%TED & 1698/93/23 & 0.21M/9K/2.3K & 120.89/96.42/98.74 \\
%News & 6069/81/155 & 0.24M/2K/3K & 38.93/26.78/19.35 \\
%Europarl & 118K/240/360 & 1.67M/3.6K/5.1K & 14.14/14.95/14.06 \\
TED & 0.21M/9K/2.3K & 120.89/96.42/98.74 \\
News & 0.24M/2K/3K & 38.93/26.78/19.35 \\
Europarl & 1.67M/3.6K/5.1K & 14.14/14.95/14.06 \\
\end{tabular}
}
\caption{Training/development/test corpora statistics: number of sentences (K stands for thousands and M for millions), and average document length (in sentences).}
\label{table:corpora}
\end{table}

%% file: table-offline-results.tex
\setlength{\tabcolsep}{1.5pt}
\begin{table*}[t]
{\small
\centering
\begin{tabular}{l | c c | c c | c c | c c | c c | c c }
& \multicolumn{6}{c}{\textbf{Integration into Encoder}} & \multicolumn{6}{|c}{\textbf{Integration into Decoder}}\\
\cline{2-13}         
&\multicolumn{2}{c}{\textbf{TED}} &\multicolumn{2}{|c}{\textbf{News}} &\multicolumn{2}{|c}{\textbf{Europarl}} & \multicolumn{2}{|c}{\textbf{TED}} &\multicolumn{2}{|c}{\textbf{News}} &\multicolumn{2}{|c}{\textbf{Europarl}}  \\  
\cline{2-13}
\textbf{Model} & \textbf{BLEU} & \textbf{Meteor} & \textbf{BLEU} & \textbf{Meteor} & \textbf{BLEU} & \textbf{Meteor} & \textbf{BLEU} & \textbf{Meteor} & \textbf{BLEU} & \textbf{Meteor} & \textbf{BLEU} & \textbf{Meteor} \\
\hline
\hline
RNNSearch & 19.24 & 40.81 & 16.51 & 36.79 & 26.26 & 44.14 & 19.24 & 40.81 & 16.51 & 36.79 & 26.26 & 44.14\\
Transformer & 23.28 & 44.17 & 22.78 & 42.19 & 28.72 & 46.22 & 23.28 & 44.17 & 22.78 & 42.19 & 28.72 & 46.22 \\
\hline
+Attention, sentence & 24.47 & \textbf{45.25} & \textbf{24.78} & 43.90 & 29.60 & 46.98 & \textbf{24.38} & 44.82 & 24.67 & 43.82 & 29.67 & \textbf{47.04} \\
\qquad\qquad\quad word & \textbf{24.55} & 44.89 & 24.55 & 43.75 & 29.63 & 46.94 & 24.27 & 44.95 & 24.23 & 43.44 & 29.68 &  46.93 \\
+H-Attention, sparse-soft & 24.23 & 44.81 & 24.76 & 44.10 & \textbf{29.72} & 47.03 & 24.19 & 44.94 & \textbf{24.67} & \textbf{43.86} & \textbf{29.69} & 46.97 \\
\qquad\qquad\qquad sparse-sparse & 24.27 & 45.07 & 24.66 & \textbf{44.18} & 29.64 & \textbf{47.04} & 24.14 & \textbf{45.32} & 24.49 & 43.49 & 29.59 & 47.02 \\
%\ \ +Hierarchical Attention & & & & & & & & & & & & \\
%\ \ \ \ \ sparsemax-softmax & 24.23 & 44.81 & 24.76 & 44.10 & \textbf{29.72} & 47.03 & 24.19 & 44.94 & \textbf{24.67} & \textbf{43.86} & \textbf{29.69} & 46.97 \\
%\ \ \ \ \ sparsemax-sparsemax & 24.27 & 45.07 & 24.66 & \textbf{44.18} & 29.64 & \textbf{47.04} & 24.14 & \textbf{45.32} & 24.49 & 43.49 & 29.59 & 47.02 \\
\hline                          
\end{tabular}
}
\caption{BLEU and Meteor scores for variants of our model and two context-agnostic baselines for offline document MT. \textbf{bold}: Best performance. All reported results for our model are significantly better than both baselines.}
\label{table:offline}
\end{table*}

%% file: table-online-results.tex
\setlength{\tabcolsep}{1.1pt}
\begin{table*}[t]
{\small
\centering
\begin{tabular}{l | c c | c c | c c | c c | c c | c c }
& \multicolumn{6}{c}{\textbf{Integration into Encoder}} & \multicolumn{6}{|c}{\textbf{Integration into Decoder}}\\
\cline{2-13}  
&\multicolumn{2}{c}{\textbf{TED}} &\multicolumn{2}{|c}{\textbf{News}} &\multicolumn{2}{|c}{\textbf{Europarl}} & \multicolumn{2}{|c}{\textbf{TED}} &\multicolumn{2}{|c}{\textbf{News}} &\multicolumn{2}{|c}{\textbf{Europarl}}  \\  
\cline{2-13}
\textbf{Model} & \textbf{BLEU} & \textbf{Meteor} & \textbf{BLEU} & \textbf{Meteor} & \textbf{BLEU} & \textbf{Meteor} & \textbf{BLEU} & \textbf{Meteor} & \textbf{BLEU} & \textbf{Meteor} & \textbf{BLEU} & \textbf{Meteor} \\
\hline
\hline
\newcite{Zhang:18} & 24.00 & 44.69 & 23.08 & 42.40 & 29.32 & 46.72 & 23.82 & 44.54 & 22.78 & 42.17 & 29.35 & 46.73\\
\newcite{Miculicich:18} & \textbf{24.58} & \textbf{45.48} & \textbf{25.03} & 44.02 & 28.60 & 46.09 & 24.39 & 45.23 & 24.38 & 43.58 & 29.58 & 46.91\\
\hline
Transformer & 23.28 & 44.17 & 22.78 & 42.19 & 28.72 & 46.22 & 23.28 & 44.17 & 22.78 & 42.19 & 28.72 & 46.22 \\
\hline
+Attention, sentence & 24.38 & 45.01 & 24.46$^\bigstar$ & 43.46$^\bigstar$ & 29.59$^\clubsuit$ & 47.02$^\clubsuit$ & 24.29$^\bigstar$ & 45.13$^\bigstar$ & {\textbf{24.75}$^\clubsuit$} & {\textbf{44.03}$^\clubsuit$} & 29.56 & 46.84\\
\qquad\qquad\quad word & 24.22 & 45.05$^\bigstar$ & 24.84$^\bigstar$ & \textbf{44.27}$^\bigstar$ & 29.67$^\clubsuit$ & 47.04$^\clubsuit$ & 24.02 & 44.79 & 24.17$^\bigstar$ & 43.53$^\bigstar$ & \textbf{29.90$^\clubsuit$} & 47.11$^\clubsuit$\\
+H-Attention, sparse-soft & 24.34 & 45.05$^\bigstar$ & 24.54$^\bigstar$ & 43.66$^\bigstar$ & \textbf{29.75$^\clubsuit$} & \textbf{47.22$^\clubsuit$} & \textbf{24.62}$^\bigstar$ & \textbf{45.32}$^\bigstar$ & 24.36$^\bigstar$ & 43.67$^\bigstar$ & 29.80$^\bigstar$ & \textbf{47.11$^\clubsuit$}\\
\qquad\qquad\qquad sparse-sparse & 24.42 & 45.38$^\bigstar$ & 24.73$^\bigstar$ & 44.06$^\bigstar$ & 29.39$^\diamondsuit$ & 46.78$^\diamondsuit$ & 24.43$^\bigstar$ & 45.10$^\bigstar$ & 24.58$^\bigstar$ & 43.75$^\bigstar$ & 29.64$^\bigstar$ & 46.94$^\bigstar$\\
%\ \ +Hierarchical Attention & & & & & & & & & & & & \\
%\ \ \ \ \ sparsemax-softmax & 24.34 & 45.05$^\clubsuit$ & 24.54$^\clubsuit$ & 43.66$^\clubsuit$ & \textbf{29.75$^\clubsuit$}$^\diamondsuit$ & \textbf{47.22$^\clubsuit$}$^\diamondsuit$ & \textbf{24.62}$^\clubsuit$ & \textbf{45.32}$^\clubsuit$ & 24.36$^\clubsuit$ & 43.67$^\clubsuit$ & 29.80$^\clubsuit$ & \textbf{47.11$^\clubsuit$}$^\diamondsuit$\\
%\ \ \ \ \ sparsemax-sparsemax & 24.42 & 45.38$^\clubsuit$ & 24.73$^\clubsuit$ & 44.06$^\clubsuit$ & 29.39$^\diamondsuit$ & 46.78$^\diamondsuit$ & 24.43$^\clubsuit$ & 45.10$^\clubsuit$ & 24.58$^\clubsuit$ & 43.75$^\clubsuit$ & 29.64$^\clubsuit$ & 46.94$^\clubsuit$\\
\hline                          
\end{tabular}
}
\caption{BLEU and Meteor scores for variants of our model and three baselines for online document MT. \textbf{bold}: Best performance. {\small{$\bigstar$}}, {\small{$\diamondsuit$}}, {\small{$\clubsuit$}}: Statistically significantly better than our implementations of \newcite{Zhang:18}, \newcite{Miculicich:18}, or both. All reported results for our model are significantly better than the Transformer.}
\label{table:online}
\end{table*}

%% file: sec6-results.tex
\subsection{Main Results}

We divide our experiments into two parts: offline and online document MT. %Here we describe the main results we obtained for both settings.

\paragraph{Offline Document MT}

From the scores of the two context-agnostic baselines in Table~\ref{table:offline}, we can see that the Transformer beats the RNNSearch model in all cases by atleast +2.5 BLEU and +2.1 Meteor scores showing that our hyperparameter choice for the Transformer is indeed effective.

For the Encoder Context integration, our Hierarchical Attention models perform the (near) best for News and Europarl datasets with +1.98 and +1 BLEU and +1.99 and +0.82  Meteor improvements with respect to the Transformer. For TED talks, however, we find the Flat Attention based models (sentence and word-level) to be the best with +1.27 BLEU and +1.08 METEOR improvements. For Decoder Context integration, we find the Hierarchical Attention to be the best in majority of the cases both in terms of BLEU and Meteor. 

\paragraph{Online Document MT}

From Table~\ref{table:online}, all our models significantly outperform the context-agnostic baseline and are significantly better than \newcite{Zhang:18} in majority cases. For Encoder Context integration, the HAN encoder \cite{Miculicich:18} is the best for TED and News datasets, however, the results are statistically insignificant with respect to our best model. For Europarl, our Hierarchical Attention model performs significantly better than \newcite{Miculicich:18} with a gain of +1.15 BLEU and +1.13 Meteor. For Decoder Context integration, our Hierachical Attention models are the winner in majority cases and our best models beat \newcite{Miculicich:18} for all datasets based upon BLEU and Meteor. \sameen{The main conclusion we draw from these results is that efficiently using the context information at hand is crucial when it comes to improving the performance of context-aware NMT. Furthermore, shorter pieces of text (e.g., the ones in Europarl) benefit more from using global context because their sentences may exhibit higher inter-dependency than those in a longer piece of text.}

\paragraph{Offline vs. Online Document MT}

Let us compare the overall results for the offline and online document MT settings. For all datasets and model variants, we find the best BLEU and Meteor scores in Tables~\ref{table:offline} and~\ref{table:online} (highlighted in bold) to be quite close to each other with those for the online setting slightly better. This is quite self-explanatory, %when we consider the way the documents in each of the datasets are produced. The datasets may be different in terms of style and level of formality,
because in essence, all of the datasets comprise of talks, speeches or commentaries, which are in fact produced in an online manner and hence we do not see drastic improvements in terms of BLEU and Meteor when conditioning on the future context. This, in our opinion, does not mean that we should never look into the future, but just that NMT models in general are highly subjective to data, and whether context-aware models benefit from future context is also dependent on that.

\subsection{Analysis}

\paragraph{Evaluation on Contrastive Pronoun Test Set}

It has been argued that evaluation metrics which quantify the overall translation quality are somewhat ill-equipped to assess how well models translate inter-sentential phenomena such as pronouns. Hence, we use a test suite of contrastive translations designed to measure accuracy of translating the English pronoun \textit{it} to its German counterparts \textit{es}, \textit{er} and \textit{sie} \cite{Mueller:18}. %The test set uses previous sentences as context and is used to evaluate our models trained on TED talks. 
We are interested to see if our global document-context models surpass the local context-aware baselines. Table~\ref{table:pronouneval} shows that not only our global-context models are quite effective but our Hierarchical Attention model is most useful when the antecedent is farther than three previous sentences. We also conclude that models for offline MT perform better when antecedent distance is greater than two.

\input{table-pronouneval.tex}

\paragraph{Subjective Evaluation}
We conduct a subjective evaluation to validate the benefit of exploiting document-level context. Three native German speakers were asked to choose the better (with ties allowed) of two translations for each of 18 documents (randomly sampled from Europarl test set). The two translations, one produced by the Transformer and the other by our Hierarchical Attention model, were evaluated in terms of: \textit{adequacy} (Which translation expresses the meaning of the source text more adequately?) and \textit{fluency} (Which text has better German?) \cite{Laubli:18}. Let \textit{a}, \textit{b} be number of ratings in favour of Transformer or our model, respectively, and \textit{t} be number of ties, then number of successes $x = b+0.5t$ and trials $n = a+b+t$. We test for statistically significant preference of our model over the Transformer by means of two-sided Sign Tests and find that our model is better than the Transformer both in terms of document-level adequacy (\textit{x} = 39, \textit{n} = 54, \textit{p} = 0.0015) and fluency (\textit{x} = 38, \textit{n} = 54, \textit{p} = 0.0038).

\paragraph{Model Complexity}
Model complexity is reported in Table~\ref{table:params}. Our context-aware models introduce only 8\% more parameters to the original Transformer model%, which is slightly less than the ones introduced by the HAN architecture \cite{Miculicich:18}
. In comparison to the Transformer, our Hierarchical Attention model is slow in training, dropping the speed by almost 50\%\footnote{DyNet implementation of \textit{sparsemax} is CPU-based and only operates on column vectors. We believe a GPU-based matrix implementation would bring the speed much closer to our Word Attention model (training: 3100, decoding: 81.38).}, but it is still almost 40\% faster than \newcite{Miculicich:18}. At decoding time, our Hierarchical Attention model is almost equivalent to \newcite{Miculicich:18} and only 13\% slower than \newcite{Zhang:18}. %The decoding speed for the Decoder-side Context models (not shown due to lack of space), drops to half which we believe to be due to the autoregressive decoding in the Transformer. Although, the overall trend is still the same.
\sameen{Hence, attending to the whole document (instead of few previous sentences) does not add to the time complexity of the model on average.}

\input{table-params.tex}

\paragraph{Qualitative Analysis}
To analyse the effect of using sparse attention at both the sentence and word-level, we looked at the attention weights computed by \textit{sparsemax}. Table~\ref{table:attmap2} shows an example where our model helped generate a correct translation of the noun ``thoughts'' (highlighted in bold). The context sentences shown in the bottom box had the highest attention weights as assigned by sparsemax. It seems that this particular attention head focuses more on phrases like ``words of sympathy'', ``support', ``symbol of hope'' which are related to the query ``thoughts''. 
Another example in Table~\ref{table:attmap} shows how our model correctly translates the pronoun ``their''. Upon looking at the words in the context sentences, it seems that this particular attention head focuses on the words related to the antecedent ``Croatia's Serbian population'' with most of the weight concentrated around neighbouring words in sentence $s^{j-1}$. It is evident from both examples that word-level sparsity is more prevalent in longer sentences in the context. The same holds for sparsity at sentence-level.

\input{table-attmap2.tex}

\input{table-attmap.tex}

%% file: table-pronouneval.tex
\setlength{\tabcolsep}{1.5pt}
\begin{table}[t]
\centering
{\small
\begin{tabular}{l|c c c c c}
\textbf{Model} & \multicolumn{5}{c}{\textbf{antecedent distance}}\\
%\cline{3-4}
& 0 & 1 & 2 & 3 & $>$3\\
\hline \hline
& \multicolumn{5}{c}{Offline document MT}\\
\hline
RNNSearch & 0.415 & 0.310 & 0.424 & 0.440 & 0.647\\
Transformer & 0.586 & 0.308 & 0.437 & 0.48 & 0.642\\
\hline
+Attention, sentence & 0.677 & 0.314 & 0.439 & 0.478 & 0.697\\
\qquad\qquad\quad word & \textbf{0.686} & \textbf{0.347} & \textbf{0.464} & \textbf{0.511} & 0.679\\
+H-Attention, sparse-soft & 0.676 & 0.308 & 0.440 & 0.480 & 0.686\\
\qquad\qquad\qquad sparse-sparse & 0.652 & 0.303 & 0.435 & 0.471 & \textbf{0.701}\\
%\ \ +Hierarchical Attention & & & & &\\
%\ \ \ \ \ sparsemax-softmax & 0.676 & 0.308 & 0.440 & 0.480 & 0.686\\
%\ \ \ \ \ sparsemax-sparsemax & 0.652 & 0.303 & 0.435 & 0.471 & \textbf{0.701}\\
\hline \hline
& \multicolumn{5}{c}{Online document MT}\\
\hline
\newcite{Zhang:18} & 0.622 & 0.321 & 0.450 & 0.485 & 0.658\\
\newcite{Miculicich:18} & 0.722 & 0.326 & 0.451 & 0.471 & 0.661\\
\hline
Transformer & 0.586 & 0.308 & 0.437 & 0.48 & 0.642\\
\hline
+Attention, sentence & \textbf{0.732} & \textbf{0.340} & \textbf{0.460} & 0.485 & 0.661\\
\qquad\qquad\quad word & 0.690 & 0.317 & 0.444 & 0.487 & 0.683\\
+H-Attention, sparse-soft & 0.692 & 0.329 & 0.446 & 0.464 & 0.656\\
\qquad\qquad\qquad sparse-sparse & 0.711 & 0.317 & 0.437 & \textbf{0.489} & \textbf{0.692}\\
%\ \ +Hierarchical Attention & & & & &\\
%\ \ \ \ \ sparsemax-softmax & 0.692 & 0.329 & 0.446 & 0.464 & 0.656\\
%\ \ \ \ \ sparsemax-sparsemax & 0.711 & 0.317 & 0.437 & \textbf{0.489} & \textbf{0.692}\\
\end{tabular}
}
\caption{Accuracy on contrastive test set with regard to antecedent distance (in sentences) on TED Talks. \sameen{Antecedent distance 0 means the pronoun occurs in the same sentence as the antecedent}.}
\label{table:pronouneval}
\end{table}

%% file: table-params.tex
\setlength{\tabcolsep}{2.5pt}
\begin{table}[t]
\centering
{\small
\begin{tabular}{l|c|c|c}
%& & & \multicolumn{2}{c}{Doc Length}\\
%\cline{4-5}
\textbf{Model} & \textbf{\#Params} & \multicolumn{2}{c}{\textbf{Speed (words/sec.)}}\\
\cline{3-4}
& & Training & Decoding\\
\hline \hline
\newcite{Zhang:18} & 59.5M & 3300 & 84.94\\
\newcite{Miculicich:18} & 54.8M & 1650 & 76.90\\
\hline
Transformer & 50M & 5100 & 86.33\\
\hline
+Attention, sentence & 53.7M & 3750 & 83.84\\
%\qquad\qquad\quad word & 53.7M & 3100 & 81.38\\
+H-Attention, sparse-soft & 54.2M & 2600 & 74.11\\
\end{tabular}
}
\caption{Model complexity for Encoder Context integration models (News-Commentary).}
\label{table:params}
\end{table}

%% file: table-attmap2.tex
\setlength{\tabcolsep}{0.5pt}
\begin{table}[t]
\centering
{\small
{\setlength{\fboxsep}{1.5pt}
\begin{tabular}{|l|}
%\multicolumn{1}{c}{Currently Translated Sentence}\\
\hline
Src: my \textbf{thoughts} are also with the victims .\\
Ref: meine \textbf{Gedanken} sind auch bei den Opfern .\\
\hline
Transformer: ich \textbf{denke} auch an die Opfer .\\
\newcite{Zhang:18}: ich \textbf{denke} auch an die Opfer .\\
\newcite{Miculicich:18}: ich \textbf{denke} auch an die Opfer . \\
\hline
Our Model: meine \textbf{Gedanken} sind auch bei den Opfern .\\
\hline
\end{tabular}
\begin{tabular}{l}
%\multicolumn{1}{c}{Context from Source Sentences. \textit{Query}: \textbf{thoughts}}\\
\hline
Head 2: Attention to related words \it{sympathy}, \it{support}, \it{hope}\\
{s$^{j-2}$}: ( FR ) Madam President , many \colorbox{magenta!2.5}{things} have already\\
been \colorbox{magenta!4.8}{said} \colorbox{magenta!1.6}{,} but I \colorbox{magenta!1.4}{would} \colorbox{magenta!2.2}{like} to \colorbox{magenta!5.4}{echo} \colorbox{magenta!4.5}{all} \colorbox{magenta!3.5}{the} \colorbox{magenta!8.3}{words} \colorbox{magenta!9.6}{of}\\
\colorbox{magenta!10.5}{sympathy} and \colorbox{magenta!6.7}{support} that have \colorbox{magenta!2.1}{already} been \colorbox{magenta!8.2}{addressed}\\
\colorbox{magenta!9.6}{to} \colorbox{magenta!4.1}{the} \colorbox{magenta!3.8}{peoples} of\colorbox{magenta!4.7}{Tunisia}and \colorbox{magenta!1.6}{Egypt} .\\
{s$^{j+4}$}: \colorbox{magenta!9.3}{it} \colorbox{magenta!10.2}{must} \colorbox{magenta!16.9}{implement} \colorbox{magenta!6.3}a \colorbox{magenta!11.5}{strong} \colorbox{magenta!8.9}{strategy} \colorbox{magenta!10.8}{towards}\\
\colorbox{magenta!8.8}{these} \colorbox{magenta!8.3}{countries} \colorbox{magenta!8.9} .\\
{s$^{j-1}$}: \colorbox{magenta!7.7}{they} \colorbox{magenta!10}{are} \colorbox{magenta!6.3}{a} \colorbox{magenta!11.2}{symbol} \colorbox{magenta!7.4}{of} \colorbox{magenta!12.4}{hope} \colorbox{magenta!9.8}{for} \colorbox{magenta!3.3}{all} \colorbox{magenta!3}{those} \colorbox{magenta!3.6}{who}\\
\colorbox{magenta!7.9}{defend} \colorbox{magenta!10}{freedom} \colorbox{magenta!6.8} .\\
%\hline
\hline
\end{tabular}
}
}
\caption{Example of noun disambiguation. Source context sentences are ordered in decreasing probability mass. The intensity of color corresponds to the attention given to a specific word before rescaling.}
\label{table:attmap2}
\end{table}

%% file: table-attmap.tex
\setlength{\tabcolsep}{0.5pt}
\begin{table}[t]
\centering
{\small
{\setlength{\fboxsep}{1.5pt}
\begin{tabular}{|l|}
%\multicolumn{1}{c}{Currently Translated Sentence}\\
\hline
Src: Croatia is \textbf{their} homeland , too .\\
Ref: Kroatien ist auch \textbf{ihre} Heimat .\\
\hline
Transformer: Kroatien ist auch \textbf{seine} Heimat .\\
Our Model: Kroatien ist auch \textbf{ihr} Heimatland .\\
\hline
\end{tabular}
\begin{tabular}{l}
%\multicolumn{1}{c}{Context from Previous Source Sentences. \textit{Query}: \textbf{their}}\\
\hline
Head 8: Attention to words related to the antecedent.\\
{s$^{j-1}$}: to name but a few , these include \colorbox{magenta!2.5}{cooperation} with \\
the Hague \colorbox{magenta!4.2}{Tribunal}, \colorbox{magenta!6.6}{efforts} \colorbox{magenta!2.2}{made} so far \colorbox{magenta!1.4}{in} \colorbox{magenta!9}{prosecuting}\\
\colorbox{magenta!10.3}{corruption}, \colorbox{magenta!5.3}{restructuring} the \colorbox{magenta!3.3}{economy} and \colorbox{magenta!1.7}{finances} and\\
greater \colorbox{magenta!6.4}{commitment} and \colorbox{magenta!3}{sincerity} {in} \colorbox{magenta!5.7}{eliminating} the\\
\colorbox{magenta!7}{obstacles} to the \colorbox{magenta!9}{return} of \colorbox{magenta!6.5}{Croatia} 's \colorbox{magenta!3.5}{Serbian} \colorbox{magenta!8.2}{population} . \\{s$^{j-4}$}: by \colorbox{magenta!2.7}{signing} a  \colorbox{magenta!7.2}{border} \colorbox{magenta!8}{arbitration} \colorbox{magenta!2.2}{agreement} with\\
its \colorbox{magenta!3.7}{neighbour} \colorbox{magenta!2.8}{Slovenia} , \colorbox{magenta!1.7}{the} \colorbox{magenta!5.4}{new} \colorbox{magenta!6.6}{Croatian} \colorbox{magenta!6.4}{Government}\\
has not only \colorbox{magenta!2.6}{eliminated} an \colorbox{magenta!5.9}{obstacle} \colorbox{magenta!1.6}{to} \colorbox{magenta!2.6}{the} \colorbox{magenta!9.4}{negotiating}\\
\colorbox{magenta!5.6}{process}, \colorbox{magenta!1.7}{but} has also \colorbox{magenta!2.4}{paved} \colorbox{magenta!1.8}{the} \colorbox{magenta!4.2}{way} for the \colorbox{magenta!7}{resolution} \\
of \colorbox{magenta!2.4}{other} \colorbox{magenta!6.6}{issues} .\\
%\hline
\hline
\end{tabular}
}
}
\caption{Example of pronoun disambiguation. Context sentences are ordered in decreasing probability mass.}
\label{table:attmap}
\end{table}

%% file: sec2-related-work.tex
The body of work in document-level MT can be broadly classified into two categories: conventional MT and neural MT.
 
\paragraph{Conventional Document-level MT}
These can further be classified into two main categories. The first, which use cache-based memories \cite{Tiedemann:10, Gong:11} and the second, which focus on specific discourse phenomema like anaphora \cite{Hardmeier:10}, lexical cohesion \cite{Xiong:13, Gong:15, Mascarell:17} and coreference \cite{Werlen:17} to name a few. Most of these approaches are, however, restrictive as they mostly involve using hand-crafted features similar to the conventional MT approaches.

\paragraph{Document-level Neural MT}
The works here can again be divided into two categories: \textit{online}---use previous context only, and \textit{offline}---use both past and future contexts. Most works fall into the former category, with those that use only a single previous sentence in the source \cite{Jean:17, Tiedemann:17, Voita:18}; one previous sentence both in source and target \cite{Bawden:17}; more than one previous source sentence \cite{Wang:17, Zhang:18}; or a few previous source and target sentences \cite{Miculicich:18}. Apart from fixing the context length, there are few works which use cache-based memories to store contextual information \cite{Tu:18, Kuang:18} and use that to improve the MT system performance. \sameen{A recent work \cite{Maruf2018} reports promising results when using the complete history for translating online conversations.}

For the offline setting, however, there is only one work that effectively uses the full document-context on both source and target-side using memory networks \cite{Maruf:18}. The debate in document-level NMT today is mostly about how much of the previous context to use and there has been no comparison between the online and offline setting except using only one previous and following sentence \cite{Voita:18}.

\sameen{\paragraph{Sparse Attention}
Sparse attention and its constrained variants have been used to address the coverage problem in NMT \cite{Malaviya2018} by limiting the amount of attention that each source word can receive. Apart from NMT, sparse attention has been shown to yield promising results for NLP tasks of textual entailment \cite{Martins:16} and summarization \cite{Niculae:17}.}

%% file: sec8-conclusions.tex
We have proposed a novel approach to hierarchical attention for context-aware NMT, based on sparse attention, which is both scalable and efficient. Experiments and evaluation on three English$\rightarrow$German datasets in offline and online document MT settings show that our approach surpasses context-agnostic and two recent context-aware baselines. The qualitative analysis shows that the sparsity at sentence-level allows our model to identify key sentences in the document context and the sparsity at word-level allows it to focus on key words in those sentences allowing for an efficient compression of memory. In future work, we plan to dig deeper on the benefits of sparse attention in terms of better interpretability of context-aware NMT models.

%% file: naaclhlt2019.bbl
\begin{thebibliography}{42}
\expandafter\ifx\csname natexlab\endcsname\relax\def\natexlab#1{#1}\fi

\bibitem[{Bahdanau et~al.(2015)Bahdanau, Cho, and Bengio}]{Bahdanau:15}
Dzmitry Bahdanau, Kyunghyun Cho, and Yoshua Bengio. 2015.
\newblock Neural machine translation by jointly learning to align and
  translate.
\newblock In \emph{Proceedings of the International Conference on Learning
  Representations}.

\bibitem[{Bawden et~al.(2018)Bawden, Sennrich, Birch, and Haddow}]{Bawden:17}
Rachel Bawden, Rico Sennrich, Alexandra Birch, and Barry Haddow. 2018.
\newblock \href {https://doi.org/10.18653/v1/N18-1118} {Evaluating discourse
  phenomena in neural machine translation}.
\newblock In \emph{Proceedings of the 2018 Conference of the North American
  Chapter of the Association for Computational Linguistics: Human Language
  Technologies, Volume 1 (Long Papers)}, pages 1304--1313. Association for
  Computational Linguistics.

\bibitem[{Cettolo et~al.(2012)Cettolo, Girardi, and Federico}]{Cettolo:12}
Mauro Cettolo, Christian Girardi, and Marcello Federico. 2012.
\newblock Wit$^3$: Web inventory of transcribed and translated talks.
\newblock In \emph{Proceedings of the 16$^{th}$ Conference of the European
  Association for Machine Translation (EAMT)}, pages 261--268, Trento, Italy.

\bibitem[{Cho et~al.(2014)Cho, {van Merrienboer}, Bahdanau, and
  Bengio}]{Cho:14}
Kyunghyun Cho, B~{van Merrienboer}, Dzmitry Bahdanau, and Yoshua Bengio. 2014.
\newblock On the properties of neural machine translation: Encoder-decoder
  approaches.
\newblock In \emph{Eighth Workshop on Syntax, Semantics and Structure in
  Statistical Translation (SSST-8)}.

\bibitem[{Clark et~al.(2011)Clark, Dyer, Lavie, and Smith}]{Clark:11}
Jonathan~H. Clark, Chris Dyer, Alon Lavie, and Noah~A. Smith. 2011.
\newblock \href {http://www.aclweb.org/anthology/P11-2031} {Better hypothesis
  testing for statistical machine translation: Controlling for optimizer
  instability}.
\newblock In \emph{Proceedings of the 49th Annual Meeting of the Association
  for Computational Linguistics: Human Language Technologies (Short Papers)},
  pages 176--181. Association for Computational Linguistics.

\bibitem[{Cohn et~al.(2016)Cohn, Hoang, Vymolova, Yao, Dyer, and
  Haffari}]{Cohn:16}
Trevor Cohn, Cong Duy~Vu Hoang, Ekaterina Vymolova, Kaisheng Yao, Chris Dyer,
  and Gholamreza Haffari. 2016.
\newblock \href {http://www.aclweb.org/anthology/N16-1102} {Incorporating
  structural alignment biases into an attentional neural translation model}.
\newblock In \emph{Proceedings of the North American Chapter of the Association
  for Computational Linguistics: Human Language Technologies}, pages 876--885.
  Association for Computational Linguistics.

\bibitem[{Dayan et~al.(2000)Dayan, Kakade, and Montague}]{Dayan:00}
Peter Dayan, Sham Kakade, and P.~Read Montague. 2000.
\newblock \href {https://doi.org/10.1038/81504} {Learning and selective
  attention}.
\newblock \emph{Nature Neuroscience}, 3:1218--1223.

\bibitem[{Gong et~al.(2011)Gong, Zhang, and Zhou}]{Gong:11}
Zhengxian Gong, Min Zhang, and Guodong Zhou. 2011.
\newblock \href {http://aclweb.org/anthology/D11-1084} {Cache-based
  document-level statistical machine translation}.
\newblock In \emph{Proceedings of the Conference on Empirical Methods in
  Natural Language Processing}, pages 909--919. Association for Computational
  Linguistics.

\bibitem[{Gong et~al.(2015)Gong, Zhang, and Zhou}]{Gong:15}
Zhengxian Gong, Min Zhang, and Guodong Zhou. 2015.
\newblock \href {https://doi.org/10.18653/v1/W15-2504} {Document-level machine
  translation evaluation with gist consistency and text cohesion}.
\newblock In \emph{Proceedings of the Second Workshop on Discourse in Machine
  Translation}, pages 33--40, Lisbon, Portugal. Association for Computational
  Linguistics.

\bibitem[{Hardmeier and Federico(2010)}]{Hardmeier:10}
Christian Hardmeier and Marcello Federico. 2010.
\newblock Modelling pronominal anaphora in statistical machine translation.
\newblock In \emph{International Workshop on Spoken Language Translation},
  pages 283--289.

\bibitem[{Jean et~al.(2017)Jean, Lauly, Firat, and Cho}]{Jean:17}
Sebastien Jean, Stanislas Lauly, Orhan Firat, and Kyunghyun Cho. 2017.
\newblock Does neural machine translation benefit from larger context?
\newblock In \emph{arXiv:1704.05135}.

\bibitem[{Kingma and Ba(2015)}]{Kingma:14}
Diederik~P. Kingma and Jimmy Ba. 2015.
\newblock Adam: {A} method for stochastic optimization.
\newblock In \emph{Proceedings of the International Conference on Learning
  Representations}.

\bibitem[{Koehn(2005)}]{Koehn:05}
Philipp Koehn. 2005.
\newblock Europarl: A parallel corpus for statistical machine translation.
\newblock In \emph{Conference Proceedings: the 10th Machine Translation
  Summit}, pages 79--86. AAMT.

\bibitem[{Koehn et~al.(2007)Koehn, Hoang, Birch, Callison-Burch, Federico,
  Bertoldi, Cowan, Shen, Moran, Zens, Dyer, Bojar, Constantin, and
  Herbst}]{Koehn:07}
Philipp Koehn, Hieu Hoang, Alexandra Birch, Chris Callison-Burch, Marcello
  Federico, Nicola Bertoldi, Brooke Cowan, Wade Shen, Christine Moran, Richard
  Zens, Chris Dyer, Ond\v{r}ej Bojar, Alexandra Constantin, and Evan Herbst.
  2007.
\newblock \href {http://www.aclweb.org/anthology/P07-2045} {Moses: Open source
  toolkit for statistical machine translation}.
\newblock In \emph{Proceedings of the 45th Annual Meeting of the ACL on
  Interactive Poster and Demonstration Sessions}, pages 177--180. Association
  for Computational Linguistics.

\bibitem[{Kuang et~al.(2018)Kuang, Xiong, Luo, and Zhou}]{Kuang:18}
Shaohui Kuang, Deyi Xiong, Weihua Luo, and Guodong Zhou. 2018.
\newblock \href {http://aclweb.org/anthology/C18-1050} {Modeling coherence for
  neural machine translation with dynamic and topic caches}.
\newblock In \emph{Proceedings of the 27th International Conference on
  Computational Linguistics}, pages 596--606. Association for Computational
  Linguistics.

\bibitem[{L{\"{a}}ubli et~al.(2018)L{\"{a}}ubli, Sennrich, and
  Volk}]{Laubli:18}
Samuel L{\"{a}}ubli, Rico Sennrich, and Martin Volk. 2018.
\newblock \href {http://aclweb.org/anthology/D18-1512} {Has machine translation
  achieved human parity? {A} case for document-level evaluation}.
\newblock In \emph{Proceedings of the Conference on Empirical Methods in
  Natural Language Processing}, pages 4791--4796. Association for Computational
  Linguistics.

\bibitem[{Lavie and Agarwal(2007)}]{Lavie:07}
Alon Lavie and Abhaya Agarwal. 2007.
\newblock \href {http://aclweb.org/anthology/W07-0734} {Meteor: An automatic
  metric for mt evaluation with high levels of correlation with human
  judgments}.
\newblock In \emph{Proceedings of the Second Workshop on Statistical Machine
  Translation}, pages 228--231. Association for Computational Linguistics.

\bibitem[{Malaviya et~al.(2018)Malaviya, Ferreira, and Martins}]{Malaviya2018}
Chaitanya Malaviya, Pedro Ferreira, and Andr{\'e} F.~T. Martins. 2018.
\newblock \href {http://aclweb.org/anthology/P18-2059} {Sparse and constrained
  attention for neural machine translation}.
\newblock In \emph{Proceedings of the 56th Annual Meeting of the Association
  for Computational Linguistics (Volume 2: Short Papers)}, volume~2, pages
  370--376.

\bibitem[{Martins and Astudillo(2016)}]{Martins:16}
Andr{\'e} F.~T. Martins and Ram{\'o}n Astudillo. 2016.
\newblock \href {http://proceedings.mlr.press/v48/martins16.html} {From softmax
  to sparsemax: A sparse model of attention and multi-label classification}.
\newblock In \emph{Proceedings of The 33rd International Conference on Machine
  Learning}, volume~48, pages 1614--1623, New York, New York, USA. PMLR.

\bibitem[{Maruf and Haffari(2018)}]{Maruf:18}
Sameen Maruf and Gholamreza Haffari. 2018.
\newblock \href {http://aclweb.org/anthology/P18-1118} {Document context neural
  machine translation with memory networks}.
\newblock In \emph{Proceedings of the 56th Annual Meeting of the Association
  for Computational Linguistics (Volume 1: Long Papers)}, pages 1275--1284.
  Association for Computational Linguistics.

\bibitem[{Maruf et~al.(2018)Maruf, Martins, and Haffari}]{Maruf2018}
Sameen Maruf, Andr{\'e} F.~T. Martins, and Gholamreza Haffari. 2018.
\newblock \href {http://www.aclweb.org/anthology/W18-6311} {Contextual neural
  model for translating bilingual multi-speaker conversations}.
\newblock In \emph{Proceedings of the Third Conference on Machine Translation:
  Research Papers}, pages 101--112, Brussels, Belgium. Association for
  Computational Linguistics.

\bibitem[{Mascarell(2017)}]{Mascarell:17}
Laura Mascarell. 2017.
\newblock \href {https://doi.org/10.18653/v1/W17-4813} {Lexical chains meet
  word embeddings in document-level statistical machine translation}.
\newblock In \emph{Proceedings of the Third Workshop on Discourse in Machine
  Translation}, pages 99--109. Association for Computational Linguistics.

\bibitem[{Miculicich et~al.(2018)Miculicich, Ram, Pappas, and
  Henderson}]{Miculicich:18}
Lesly Miculicich, Dhananjay Ram, Nikolaos Pappas, and James Henderson. 2018.
\newblock \href {http://aclweb.org/anthology/D18-1325} {Document-level neural
  machine translation with hierarchical attention networks}.
\newblock In \emph{Proceedings of the 2018 Conference on Empirical Methods in
  Natural Language Processing}, pages 2947--2954.

\bibitem[{Miculicich~Werlen and Popescu-Belis(2017)}]{Werlen:17}
Lesly Miculicich~Werlen and Andrei Popescu-Belis. 2017.
\newblock \href {https://doi.org/10.18653/v1/W17-1505} {Using coreference links
  to improve spanish-to-english machine translation}.
\newblock In \emph{Proceedings of the 2nd Workshop on Coreference Resolution
  Beyond OntoNotes (CORBON 2017)}, pages 30--40, Valencia, Spain. Association
  for Computational Linguistics.

\bibitem[{M{\"u}ller et~al.(2018)M{\"u}ller, Rios, Voita, and
  Sennrich}]{Mueller:18}
Mathias M{\"u}ller, Annette Rios, Elena Voita, and Rico Sennrich. 2018.
\newblock \href {http://www.aclweb.org/anthology/W18-6307} {A large-scale test
  set for the evaluation of context-aware pronoun translation in neural machine
  translation}.
\newblock In \emph{Proceedings of the Third Conference on Machine Translation:
  Research Papers}, pages 61--72, Belgium, Brussels. Association for
  Computational Linguistics.

\bibitem[{Nallapati et~al.(2016)Nallapati, Zhou, dos Santos,
  G{\"{u}}l{\c{c}}ehre, and Xiang}]{Nallapati:16}
Ramesh Nallapati, Bowen Zhou, C{\'{\i}}cero~Nogueira dos Santos, {\c{C}}aglar
  G{\"{u}}l{\c{c}}ehre, and Bing Xiang. 2016.
\newblock \href {http://www.aclweb.org/anthology/K16-1028} {Abstractive text
  summarization using sequence-to-sequence {RNN}s and beyond}.
\newblock In \emph{Proceedings of Conference on Natural Language Learning},
  pages 280--290. Association for Computational Linguistics.

\bibitem[{Neubig et~al.(2017)Neubig, Dyer, Goldberg, Matthews, Ammar,
  Anastasopoulos, Ballesteros, Chiang, Clothiaux, Cohn, Duh, Faruqui, Gan,
  Garrette, Ji, Kong, Kuncoro, Kumar, Malaviya, Michel, Oda, Richardson,
  Saphra, Swayamdipta, and Yin}]{dynet}
Graham Neubig, Chris Dyer, Yoav Goldberg, Austin Matthews, Waleed Ammar,
  Antonios Anastasopoulos, Miguel Ballesteros, David Chiang, Daniel Clothiaux,
  Trevor Cohn, Kevin Duh, Manaal Faruqui, Cynthia Gan, Dan Garrette, Yangfeng
  Ji, Lingpeng Kong, Adhiguna Kuncoro, Gaurav Kumar, Chaitanya Malaviya, Paul
  Michel, Yusuke Oda, Matthew Richardson, Naomi Saphra, Swabha Swayamdipta, and
  Pengcheng Yin. 2017.
\newblock Dynet: The dynamic neural network toolkit.
\newblock \emph{arXiv preprint arXiv:1701.03980}.

\bibitem[{Niculae and Blondel(2017)}]{Niculae:17}
Vlad Niculae and Mathieu Blondel. 2017.
\newblock \href
  {http://papers.nips.cc/paper/6926-a-regularized-framework-for-sparse-and-structured-neural-attention.pdf}
  {A regularized framework for sparse and structured neural attention}.
\newblock In I.~Guyon, U.~V. Luxburg, S.~Bengio, H.~Wallach, R.~Fergus,
  S.~Vishwanathan, and R.~Garnett, editors, \emph{Advances in Neural
  Information Processing Systems 30}, pages 3338--3348. Curran Associates, Inc.

\bibitem[{Papineni et~al.(2002)Papineni, Roukos, Ward, and Zhu}]{Papineni:02}
Kishore Papineni, Salim Roukos, Todd Ward, and Wei-Jing Zhu. 2002.
\newblock \href {http://aclweb.org/anthology/P02-1040} {{BLEU}: A method for
  automatic evaluation of machine translation}.
\newblock In \emph{Proceedings of the 40th Annual Meeting on Association for
  Computational Linguistics}, pages 311--318. Association for Computational
  Linguistics.

\bibitem[{Sennrich et~al.(2016)Sennrich, Haddow, and Birch}]{Sennrich:16}
Rico Sennrich, Barry Haddow, and Alexandra Birch. 2016.
\newblock \href {http://www.aclweb.org/anthology/P16-1162} {Neural machine
  translation of rare words with subword units}.
\newblock In \emph{Proceedings of the 54$^{th}$ Annual Meeting of the
  Association for Computational Linguistics}, pages 1715--1725.

\bibitem[{Sutskever et~al.(2014)Sutskever, Vinyals, and Le}]{Sutskever:14}
Ilya Sutskever, Oriol Vinyals, and Quoc~V. Le. 2014.
\newblock \href
  {http://papers.nips.cc/paper/5346-sequence-to-sequence-learning-with-neural-networks.pdf}
  {Sequence to sequence learning with neural networks}.
\newblock In \emph{Proceedings of the 27th International Conference on Neural
  Information Processing Systems}, pages 3104--3112. MIT Press.

\bibitem[{Tiedemann(2010)}]{Tiedemann:10}
J\"{o}rg Tiedemann. 2010.
\newblock \href {http://www.aclweb.org/anthology/W10-2602} {Context adaptation
  in statistical machine translation using models with exponentially decaying
  cache}.
\newblock In \emph{Proceedings of the 2010 Workshop on Domain Adaptation for
  Natural Language Processing}, DANLP 2010, pages 8--15, Stroudsburg, PA, USA.
  Association for Computational Linguistics.

\bibitem[{Tiedemann(2012)}]{Tiedemann:12}
J{\"o}rg Tiedemann. 2012.
\newblock Parallel data, tools and interfaces in {OPUS}.
\newblock In \emph{Proceedings of the 8th International Conference on Language
  Resources and Evaluation (LREC'12)}, Istanbul, Turkey. European Language
  Resources Association (ELRA).

\bibitem[{Tiedemann and Scherrer(2017)}]{Tiedemann:17}
J{\"o}rg Tiedemann and Yves Scherrer. 2017.
\newblock \href {https://doi.org/10.18653/v1/W17-4811} {Neural machine
  translation with extended context}.
\newblock In \emph{Proceedings of the Third Workshop on Discourse in Machine
  Translation}, pages 82--92. Association for Computational Linguistics.

\bibitem[{Tu et~al.(2018)Tu, Liu, Shi, and Zhang}]{Tu:18}
Zhaopeng Tu, Yang Liu, Shuming Shi, and Tong Zhang. 2018.
\newblock \href {http://aclweb.org/anthology/Q18-1029} {Learning to remember
  translation history with a continuous cache}.
\newblock \emph{Transactions of the Association for Computational Linguistics},
  6:407--420.

\bibitem[{Vaswani et~al.(2017)Vaswani, Shazeer, Parmar, Uszkoreit, Jones,
  Gomez, Kaiser, and Polosukhin}]{Vaswani:17}
Ashish Vaswani, Noam Shazeer, Niki Parmar, Jakob Uszkoreit, Llion Jones,
  Aidan~N Gomez, {\L}ukasz Kaiser, and Illia Polosukhin. 2017.
\newblock \href
  {http://papers.nips.cc/paper/7181-attention-is-all-you-need.pdf} {Attention
  is all you need}.
\newblock In \emph{Advances in Neural Information Processing Systems 30}, pages
  5998--6008. Curran Associates, Inc.

\bibitem[{Voita et~al.(2018)Voita, Serdyukov, Sennrich, and Titov}]{Voita:18}
Elena Voita, Pavel Serdyukov, Rico Sennrich, and Ivan Titov. 2018.
\newblock \href {http://aclweb.org/anthology/P18-1117} {Context-aware neural
  machine translation learns anaphora resolution}.
\newblock In \emph{Proceedings of the 56th Annual Meeting of the Association
  for Computational Linguistics (Volume 1: Long Papers)}, pages 1264--1274.
  Association for Computational Linguistics.

\bibitem[{Wang et~al.(2017)Wang, Tu, Way, and Liu}]{Wang:17}
Longyue Wang, Zhaopeng Tu, Andy Way, and Qun Liu. 2017.
\newblock \href {http://aclweb.org/anthology/D17-1301} {Exploiting
  cross-sentence context for neural machine translation}.
\newblock In \emph{Proceedings of the Conference on Empirical Methods in
  Natural Language Processing}, pages 2816--2821. Association for Computational
  Linguistics.

\bibitem[{Xia et~al.(2018)Xia, Tan, Tian, Qin, Yu, and Liu}]{Xia:18}
Yingce Xia, Xu~Tan, Fei Tian, Tao Qin, Nenghai Yu, and Tie{-}Yan Liu. 2018.
\newblock \href {http://proceedings.mlr.press/v80/xia18a.html} {Model-level
  dual learning}.
\newblock In \emph{Proceedings of the 35th International Conference on Machine
  Learning}, pages 5379--5388.

\bibitem[{Xiong et~al.(2013)Xiong, Ding, Zhang, and Tan}]{Xiong:13}
Deyi Xiong, Yang Ding, Min Zhang, and Chew~Lim Tan. 2013.
\newblock \href {http://aclweb.org/anthology/D/D13/D13-1163.pdf} {Lexical chain
  based cohesion models for document-level statistical machine translation}.
\newblock In \emph{Proceedings of the 2013 Conference on Empirical Methods in
  Natural Language Processing}, pages 1563--1573. Association for Computational
  Linguistics.

\bibitem[{Yang et~al.(2016)Yang, Yang, Dyer, He, Smola, and Hovy}]{Yang:16}
Zichao Yang, Diyi Yang, Chris Dyer, Xiaodong He, Alex Smola, and Eduard Hovy.
  2016.
\newblock \href {https://doi.org/10.18653/v1/N16-1174} {Hierarchical attention
  networks for document classification}.
\newblock In \emph{Proceedings of the 2016 Conference of the North American
  Chapter of the Association for Computational Linguistics: Human Language
  Technologies}, pages 1480--1489. Association for Computational Linguistics.

\bibitem[{Zhang et~al.(2018)Zhang, Luan, Sun, Zhai, Xu, Zhang, and
  Liu}]{Zhang:18}
Jiacheng Zhang, Huanbo Luan, Maosong Sun, Feifei Zhai, Jingfang Xu, Min Zhang,
  and Yang Liu. 2018.
\newblock \href {http://aclweb.org/anthology/D18-1049} {Improving the
  transformer translation model with document-level context}.
\newblock In \emph{Proceedings of the Conference on Empirical Methods in
  Natural Language Processing}, pages 533--542. Association for Computational
  Linguistics.

\end{thebibliography}
